\documentclass{article}

\usepackage[english]{babel}
\usepackage[numbers]{natbib}  

\usepackage[letterpaper,top=2cm,bottom=2cm,left=3cm,right=3cm,marginparwidth=1.75cm]{geometry}
\usepackage{xcolor}
\usepackage{amsmath,amssymb,amsfonts}
\usepackage{booktabs}
\usepackage{amsmath}
\usepackage{graphicx}
\usepackage[colorlinks=true, allcolors=blue]{hyperref}

\title{ControlFill: Spatially Adjustable Image Inpainting from Prompt Learning}
\author{Boseong Jeon\thanks{AI Model Team at Samsung Research}}
\begin{document}
\date{}
\maketitle

\begin{abstract}
In this report, I present an inpainting framework named \textit{ControlFill}, which involves training two distinct prompts: one for generating plausible objects within a designated mask (\textit{creation}) and another for filling the region by extending the background (\textit{removal}).
During the inference stage, these learned embeddings guide a diffusion network that operates without requiring heavy text encoders.
By adjusting the relative significance of the two prompts and employing classifier-free guidance, users can control the intensity of removal or creation.
Furthermore, I introduce a method to spatially vary the intensity of guidance by assigning different scales to individual pixels.
\end{abstract}

\section{Introduction}
\subsection{Image Inpainting with Diffusion Priors}

Image inpainting enables users to selectively alter specific regions of an image. Deep learning has driven significant breakthroughs in this area, particularly with the advent of Generative Adversarial Networks (GANs) \cite{suvorov2022resolution}. However, these networks often struggle to generate novel objects or produce original content from the existing context, frequently yielding blurry and unrealistic outputs due to the "regression-to-mean" phenomenon.
In response, recent developments in text-to-image diffusion models have significantly advanced the field \cite{saharia2022palette}. Prominent examples, such as SD-Inpainting \cite{sdinpainting} and ControlNet-Inpainting \cite{controlnetinpainting}, leverage the extensive capabilities of diffusion priors. This model has become an industry benchmark, as noted by sources like Automatic1111 \cite{automatic1111}. These inpainting approaches generate more convincing and visually appealing images compared to those produced by GAN-based methods.

\begin{figure*}[t] 
\centering
\includegraphics[width=0.6\textwidth]{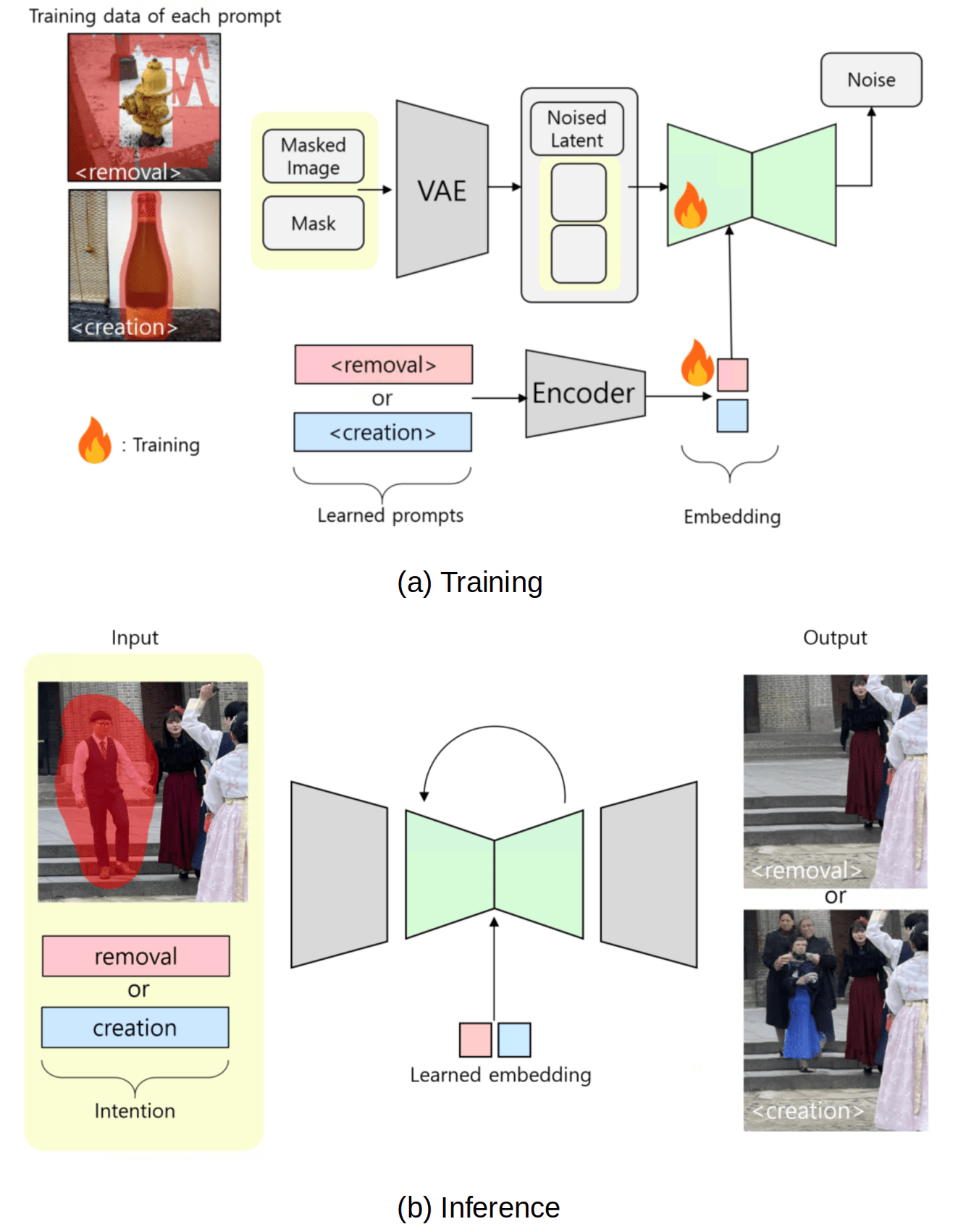}
\caption{Training and inference phases of ControlFill. (a) The model and prompts are trained to learn the two concepts of removal and creation. (b) During inference, these concepts can be applied by adjusting positive and negative prompts, without requiring a text encoder.}
\label{fig_system}
\end{figure*}

\begin{figure*}[t] 
\centering
\includegraphics[width=0.95\textwidth]{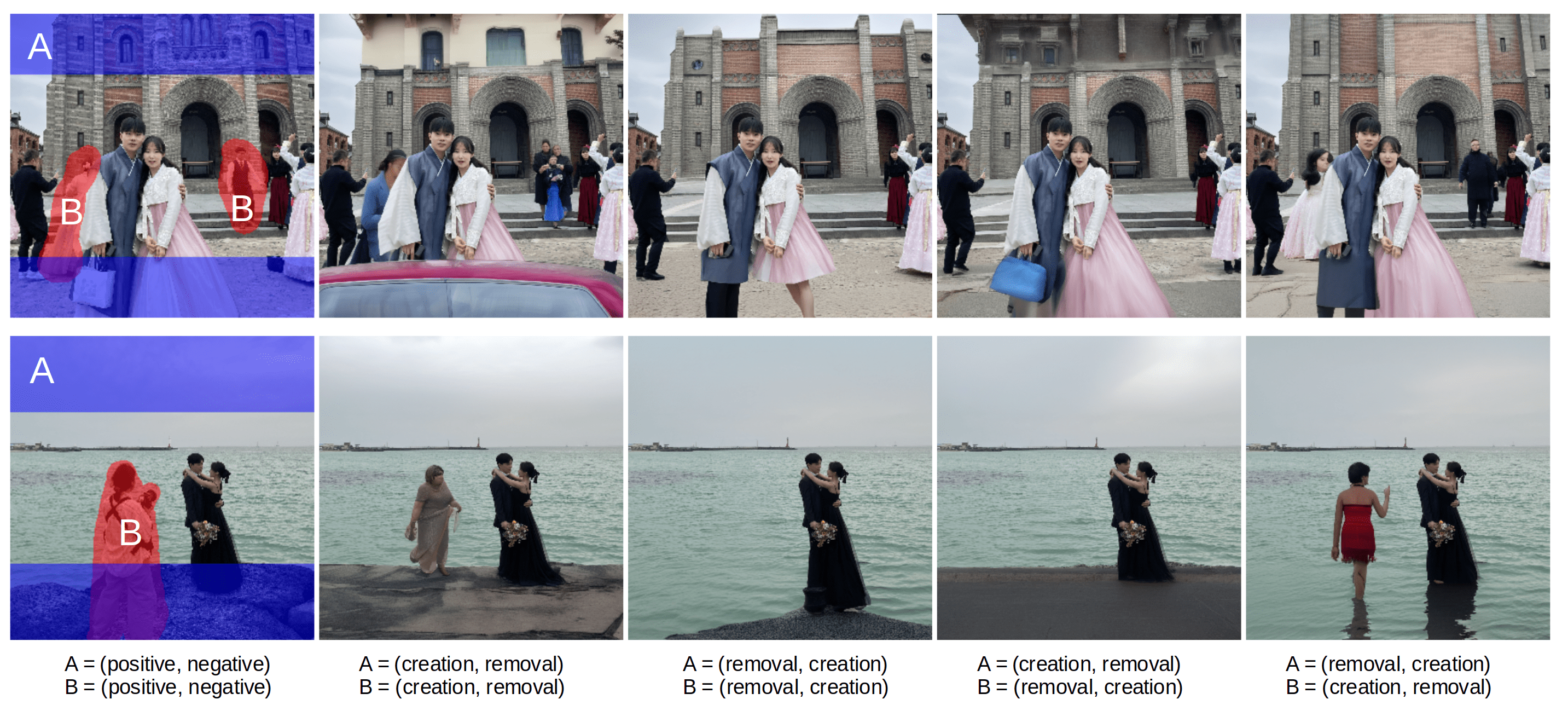}
\caption{ControlFill can reflect two user intentions (creation/removal) into individual mask regions (A and B in this example) within a single inference without text encoders. The tuples in the bottom descriptions denote positive and negative intentions, respectively.}
\label{fig_thumbnail}
\end{figure*}

Despite the innovations of diffusion-based inpainting, the approach can sometimes yield unexpected results if the prompts are not well-defined. For example, when users attempt to remove an unwanted object by masking it, the diffusion process might inadvertently introduce another object into the masked area. Conversely, in cases where users seek texture-rich and diverse content, the network might generate overly simplistic and barren scenes.

Many users find themselves needing to use multiple text prompts for seemingly straightforward tasks like object removal, as highlighted in various studies \cite{sdremoval}. For removal tasks, users may try prompts like "background," "scenery," or "no objects," yet these do not always ensure clean removal without unwanted objects. For instance, as illustrated in Figure \ref{fig_prompt}, describing the unmasked region (e.g., a "chair") can be more effective than using prompts like "background, empty."
In fact, many users prefer inpainting without any prompts at all, as discussed in community forums \cite{promptless}. It appears that crafting multiple prompts is challenging, even for seemingly simple tasks like object removal. Furthermore, integrating a text encoder into the network solely for inpainting purposes can be inefficient, particularly when memory constraints are a concern in applications like on-device processing.

\subsection{Scope and Objectives}
\label{sec:Objective}
This paper focuses on two primary applications of inpainting: (1) eliminating undesired objects (\textit{removal}) and (2) generating plausible content to create more lively images (\textit{creation}). A typical use case for removal is an AI eraser, while creation is more related to image extension \cite{wu2022nuwa} or adding relevant objects (e.g., placing more dishes on a dining table).

In this context, I aim to provide users with a seamless inpainting experience that allows straightforward control over these two tasks without requiring complex prompt engineering. Furthermore, I enable pixel-specific control rather than applying a uniform scale across the entire image. This simplifies inference by removing the need for multiple inferences when different intentions must be reflected in a single image. Additionally, I prioritize efficiency in memory usage and computational speed to facilitate on-device applications by eliminating text encoders from the inference pipeline.
\section{Related Works}
\subsection{Diffusion Models}
Diffusion models (DMs) \cite{ho2020denoising} have garnered significant attention recently due to their stability and superior performance in image synthesis compared to GANs \cite{dhariwal2021diffusion}. These models learn the data distribution by reversing a Markov noising process. Starting with a clean image $x_0$, the diffusion process sequentially adds noise at each step $t$, resulting in a series of noisy latents $x_t$. The model is then trained to reconstruct the clean image $x_0$ from $x_t$ in the reverse process. DMs have demonstrated impressive results in various tasks, such as unconditional image generation \cite{ho2022cascaded}, text-to-image generation \cite{ramesh2022hierarchical}, image inpainting \cite{avrahami2023blended, lugmayr2022repaint}, and image restoration \cite{yu2024scaling}.

\subsection{Inpainting with Diffusion Models}
Inpainting DMs \cite{avrahami2023blended, lugmayr2022repaint} are typically trained from text-to-image models to leverage the generative prior. These models support text-guided inpainting by incorporating text encoders such as CLIP \cite{radford2021learning}, enabling more editability according to user intentions.

However, diffusion-based inpainting models may not be the best approach for simple use cases where users intend to either remove objects or fill in areas consistent with the background.

\textbf{Memory and Computation Constraints:} Requiring text encoders leads to increased memory consumption and computation, which can be a bottleneck for on-device applications. For example, Stable Diffusion 1.5 uses CLIP VIT-L/14 with 123 million parameters (14 percent of UNet), and Imagen \cite{saharia2022photorealistic} uses T5-XXL with 4.6 billion parameters.

\begin{figure*}[t] 
\centering
\includegraphics[width=0.85\textwidth]{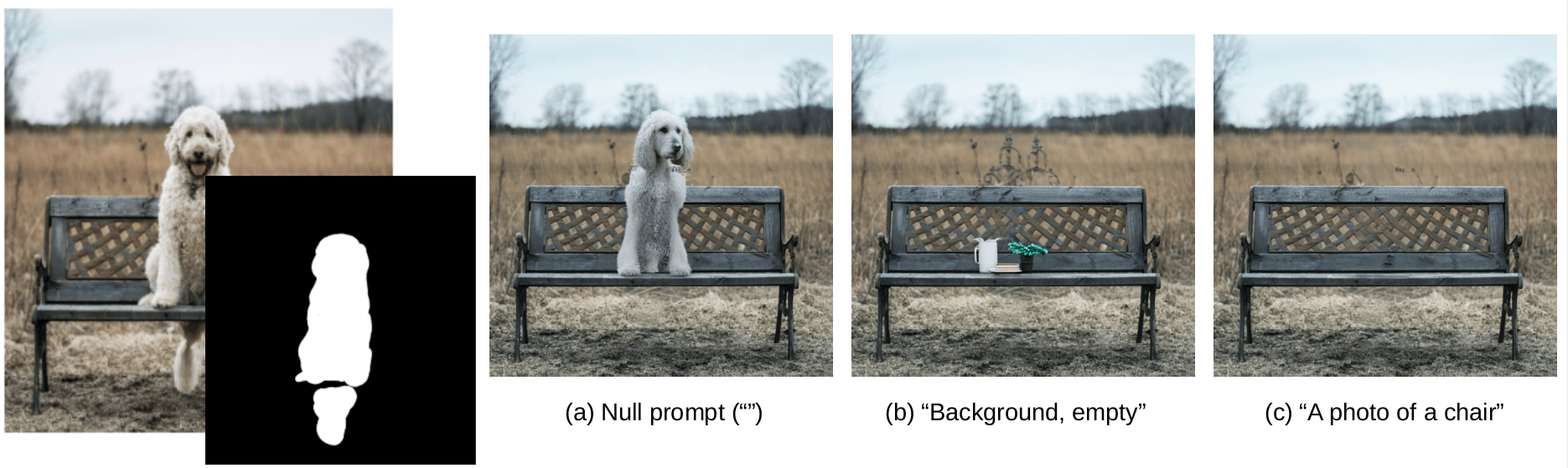}
\caption{Generated images by inpainting with diffusion priors when different prompts given. In this case, users try to remove the dog on the chair. Using fixed prompts such as \textit{empty, background} might not work for all cases, leading to unwanted objects on the chair. The best prompt in this case is related with describing the nearest unmasked region.}
\label{fig_prompt}
\end{figure*}

\textbf{Challenges in Prompt Engineering:} Creating effective prompts is not as straightforward as it seems for these tasks. Although DMs are capable of realistic image generation, they can produce unwanted random objects in the editing areas, as noted in academic papers \cite{suvorov2022resolution, wang2023towards} and by the community \cite{unwantedobject}. To address this issue, some users who want to remove objects rely on prompts like "background, empty" to avoid unwanted objects. However, as shown in Figure \ref{fig_prompt}, I found empirically that specific descriptions of an entity can sometimes be more effective than static prompts. This indicates that prompt engineering can be more challenging for simple user intentions (such as object removal) than for filling a masked region with a specific object (e.g., a dog or human) described in text prompts.

\section{Related Works}
\subsection{Diffusion Models}
Diffusion models (DMs) \cite{ho2020denoising} have garnered significant attention recently due to their stability and superior performance in image synthesis compared to GANs \cite{dhariwal2021diffusion}. These models learn the data distribution by reversing a Markov noising process. Starting with a clean image $x_0$, the diffusion process sequentially adds noise at each step $t$, resulting in a series of noisy latents $x_t$. The model is then trained to reconstruct the clean image $x_0$ from $x_t$ in the reverse process. DMs have demonstrated impressive results in various tasks, such as unconditional image generation \cite{ho2022cascaded}, text-to-image generation \cite{ramesh2022hierarchical}, image inpainting \cite{avrahami2023blended, lugmayr2022repaint}, and image restoration \cite{yu2024scaling}.

\subsection{Inpainting with Diffusion Models}
Inpainting DMs \cite{avrahami2023blended, lugmayr2022repaint} are typically trained from text-to-image models to leverage the generative prior. These models support text-guided inpainting by incorporating text encoders such as CLIP \cite{radford2021learning}, enabling more editability according to user intentions.

However, diffusion-based inpainting models may not be the best approach for simple use cases where users intend to either remove objects or fill in areas consistent with the background.

\textbf{Memory and Computation Constraints:} Requiring text encoders leads to increased memory consumption and computation, which can be a bottleneck for on-device applications. For example, Stable Diffusion 1.5 uses CLIP VIT-L/14 with 123 million parameters (14 percent of UNet), and Imagen \cite{saharia2022photorealistic} uses T5-XXL with 4.6 billion parameters.

\textbf{Challenges in Prompt Engineering:} Creating effective prompts is not as straightforward as it seems for these tasks. Although DMs are capable of realistic image generation, they can produce unwanted random objects in the editing areas, as noted in academic papers \cite{suvorov2022resolution, wang2023towards} and by the community \cite{unwantedobject}. To address this issue, some users who want to remove objects rely on prompts like "background, empty" to avoid unwanted objects. However, as shown in Figure \ref{fig_prompt}, I found empirically that specific descriptions of an entity can sometimes be more effective than static prompts. This indicates that prompt engineering can be more challenging for simple user intentions (such as object removal) than for filling a masked region with a specific object (e.g., a dog or human) described in text prompts.

\subsection{Learning Abstract Concepts for Diffusion Models}
Several efforts have been made to enable DMs to infer concepts from representative examples rather than relying on manual descriptions.
For example, textual inversion \cite{gal2022image} helps capture and represent concepts when explicit prompts or direct instructions are not readily available, such as personal objects or artistic styles, by describing them using new "words" in the embedding space of pre-trained text-to-image models.
Similarly, Dreambooth \cite{ruiz2023dreambooth} fine-tunes a pre-trained text-to-image model (UNet) to associate a unique identifier with a specific subject. By embedding the subject within the model's output domain, this identifier can then be used to generate entirely new, photorealistic images of the subject in various scenes and contexts.

The authors of \cite{zhuang2023task} explored learning inpainting concepts such as filling with background and objects using methods similar to textual inversion. They argued that optimizing a prompt associated with a random mask could fill the mask without introducing unwanted objects. However, in practice, their mask generation approach was found to be ineffective for object removal. This will be analyzed later. Another limitation of their method is that it is not readily extensible to high-resolution image inpainting, such as SDXL, which has two heavy text encoders and requires more than 40GB of VRAM. In fact, I found that the authors did not address whether their methods can be readily applied to SDXL \cite{powerpaint}.

In this work, I adopt a similar approach to \cite{zhuang2023task} but enhance removal performance by proposing an improved mask generation method. Additionally, I develop a concept-learning method that does not require text encoders by optimizing the direct conditional inputs to UNet. I fine-tune the UNet and train conditional embeddings to understand the two concepts outlined in Section \ref{sec:Objective}.

\subsection{Classifier-Free Guidance}
Classifier-free guidance (CFG) is an approach proposed to enhance the quality and diversity of generated samples in generative models. This technique involves updating the score function to blend the gradients from both a conditional model and an unconditional model. By eliminating the need for a separate classifier, this method simplifies the training process and increases flexibility. The score update rules, as proposed, optimize the balance between the conditional and unconditional components, resulting in more accurate and diverse outputs.

In practice, given a condition $y=(y_{neg}, y_{pos})$ where $y_{neg}$ is the negative prompt and $y_{pos}$ is the positive prompt, I use the following score update:
\begin{equation}
\tilde{\epsilon}{\theta}(x, y) = (1 + w) \epsilon{\theta}(x, y_{pos}) - w \epsilon_{\theta}(x, y_{neg})
\label{eqn_cfg}
\end{equation}
where $w \in \mathbb{R}$ is a scalar weight greater than one.

The above classifier-free guidance cannot be directly applied in cases where the effect of prompts should vary spatially, as it uses only a single global weight $w$. In more recent work \cite{shen2024rethinking}, the authors proposed Semantic-aware Classifier-Free Guidance (S-CFG), which allows for different guidance levels for distinct semantic units within an image, thereby enforcing uniformity of text guidance.
Although the purposes differ, I adopt this approach to reflect two different intentions (removal and creation) for each region when multiple masks are provided.

\section{Methodology}
\subsection{Inpainting Diffusion Model}
My diffusion model, ControlFill, is constructed based on the well-trained text-to-image diffusion model, specifically, Stable Diffusion, which includes both forward and reverse processes \cite{podell2023sdxl}. In the forward process, noise is added to the latent \(z_0\) of a clean image \(x_0\) in a closed form:
\begin{equation}
z_t = \sqrt{\alpha_t} z_0 + \sqrt{1 - \alpha_t} \epsilon, \quad \epsilon \sim \mathcal{N}(0, I),
\end{equation}
where \(z_t\) is the noisy image at timestep \(t\), and \(\alpha_t\) denotes the corresponding noise level. In the reverse process, a neural network parameterized by \(\theta\), denoted as \(\epsilon_\theta\), is optimized to predict the added noise \(\epsilon_t\). This allows for the generation of images by denoising step by step from Gaussian noise. A classical diffusion model is typically optimized by:
\begin{equation}
\mathcal{L} = \mathbb{E}_{z_0, t, \epsilon_t} \left\| \epsilon_t - \epsilon_\theta(z_t, t) \right\|_2^2.
\end{equation}
\subsection{Learning Task Prompts for Inpainting}
\label{subsection_learning_task_prompts_for_inpainting}

Now, I focus on how to make my model understand the two inpainting concepts: removal and creation. Instead of using hand-crafted prompts, I determine these prompts numerically to best describe the two concepts, similar to textual inversion. The two prompts are denoted as conditional variables $y_{c}$ and $y_{r}$.

To determine $(y_{c}, y_{r})$, I utilize the training dataset to teach the model the concepts of filling masks in both creation and removal modes and to identify general prompts that are independent of specific images.

For this purpose, I optimize the UNet $\epsilon_\theta(z'_t, y, t)$ and the embeddings of the conditional variables $y_{c}$ and $y_{r}$ simultaneously. While I could train the text embedding lookups of the newly added words as in textual inversion \cite{gal2022image}, I found that directly training the conditional input $y$ is also effective without requiring additional GPU memory for text encoders. This approach is particularly beneficial when applying this method to large models such as SDXL \cite{podell2023sdxl}, which have two large text encoders. Thus, this optimization formulation is more extensible to other foundation models than that of \cite{zhuang2023task}.

\begin{figure*}[h!] 
\centering
\includegraphics[width=0.95\textwidth]{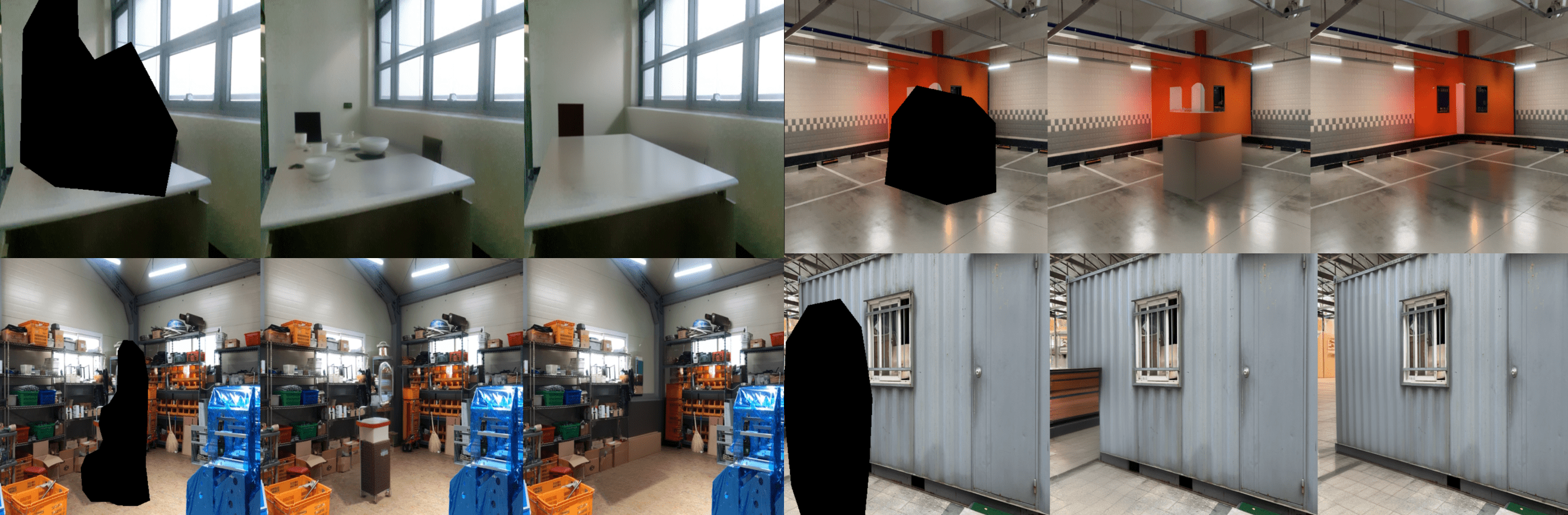}
\caption{Object removal performance depending on the mask generation method, comparing with \cite{zhuang2023task}. The left shows results from the model trained with a random mask for the removal concept, while the right demonstrates my proposed mask generation method, which strictly avoids foreground objects.}
\label{fig_abalation}
\end{figure*}

For the creation conditional embedding $y_{c}$, I provide a dataset where foreground objects are inside the mask, prompting the model to generate coherent objects by referencing the surrounding unmasked areas (background). This is illustrated in the lower part of the dataset in Figure 1-(a). To demonstrate the effectiveness of my mask generation method (which strictly avoids foreground objects) compared to \cite{zhuang2023task}, I performed an ablation study on a large mask scenario shown in Figure \ref{fig_abalation}. The baseline model was trained with random mask generation following \cite{zhuang2023task}. The results showed that my proposed method produced fewer objects inside the mask.

Conversely, for the removal condition $y_{r}$, I mask only the background while ensuring no foreground objects are present, as depicted on the top left side of the fed data in Figure 1-(a). For background masking, I adopted similar random masks as used in \cite{zhao2021large}, while excluding the region corresponding to the foreground objects. 
After training the UNet alongside the two concepts $(y_c, y_r)$, I can remove the text encoders if they were used to compute $(y_c, y_r)$.

\subsection{Spatially Varying Classifier-Free Guidance}
\label{section: Spatially Varying Classifier Free Guidance}
With the learned conditions $(y_c, y_r)$, I can use CFG to fill the masks in an image using either creation or removal mode. If multiple masks are present and belong to different conditions, I should perform two separate inferences. For example, if there are two masks with different conditions, I perform inference for one of the masks in removal mode, where $y_c$ is negative and $y_r$ is positive in (\ref{eqn_cfg}). In this section, I show that a single inference step is sufficient to reflect multiple intentions by leveraging an approach similar to S-CFG \cite{shen2024rethinking}.

Let us assume that each pixel of the mask can have two user intentions (creation and removal).
To reflect these intentions, I define a ternary mask matrix $M$ as follows:
\begin{equation}
(M)_{ij} =     
\begin{cases}
+1, & \text{creation mask} \\
-1, & \text{removal mask}\\
0, & \text{not applied} \\
\end{cases}
\label{eqn_ternary}
\end{equation}
If the latent diffusion model \cite{rombach2022high} is used as the diffusion prior, the mask matrix $M$ is downsampled to the corresponding latent size (e.g., for SDXL, 128 × 128 latent size) before being fed into the input $z'$.

For a given positive weight scalar $w$, I modify (\ref{eqn_cfg}) as follows:
\begin{equation}
\tilde{\epsilon}_{\theta}(x, y) = wM \epsilon_{\theta}(x, y_{c}) + (I - wM) \epsilon_{\theta}(x, y_{r})
\label{eqn_cfg_modify}
\end{equation}
From the above equation, the latent pixels inside the creation region have $\epsilon_{\theta}(x, y_{c})$ as a positive prompt and $\epsilon_{\theta}(x, y_{r})$ as a negative prompt, and vice versa for pixels inside the removal zone.

Unlike state-of-the-art inpainting tools such as Adobe Firefly and Google Vertex AI, which require multiple inference phases for different masking regions, my approach introduces a significant enhancement: transforming the guidance intensity from a scalar to a tensor. This modification allows for variable applications across latent features. Consequently, users can achieve diverse inpainting intentions in a single inference step. As demonstrated in Figure 1, my method effectively captures and reflects varied user intentions in just one inference.

\section{Experiments}

\subsection{Implementation}
I used the SDXL \cite{podell2023sdxl} Text-to-Image (T2I) structure.

To train ControlFill, I employed the SDXL Text-to-Image model with LoRA adaptation \cite{hu2021lora} for suitability in on-device applications. Typically, different adapters exist on top of a single foundation model to reduce the total model size. The LoRA layers were attached to all modules with a rank of 128. For training, I set the batch size to 64 and the learning rate to 1E-4, decaying it by a factor of 1E-2. AdamW \cite{loshchilov2017decoupled} was used as the optimizer.

My training process consists of two stages. In the first stage, I train only the UNet to learn the inpainting task (filling missing parts using unmasked regions) for 100K steps. In the second stage, I train both the UNet and the condition embeddings $y_{c}$ and $y_{r}$ for a shorter period of 30K steps.

\begin{figure*}[t] 
\centering
\includegraphics[width=0.85\textwidth]{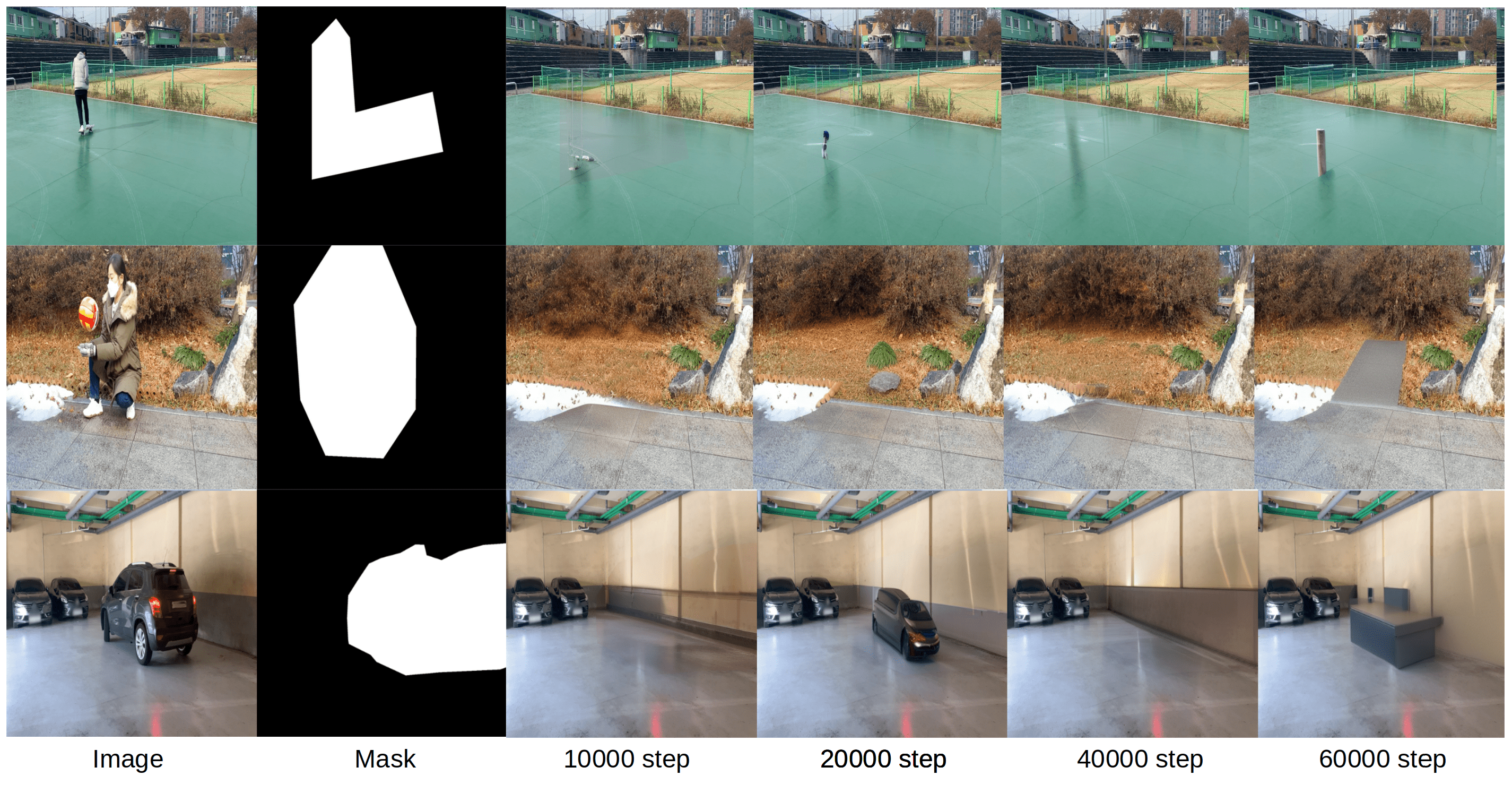}
\caption{Validation images during training. The inpainting output is pasted into the masked region to observe harmonization. As training progresses, the inpainting performance improves in terms of color matching with the unmasked region. However, object removal performance can degrade.}
\label{fig_trainig_worse}
\end{figure*}

I empirically found that training both the UNet and the embeddings $(y_c, y_r)$ for an extended period results in decreased controllability of prompts. For example, as shown in Figure \ref{fig_trainig_worse}, the \textit{removal} prompt works well in the early training steps (10K to 40K) but loses effectiveness as training progresses. I conjecture that this behavior stems from inaccuracies in the training dataset associated with each intention. 
To mitigate this issue, I train the condition embeddings only for a short period. The conditional embeddings were initialized as the text encoder output of an empty string. For the training dataset, I used commercially available Shutterstock images.

\subsection{Removal Performance}
In this section, I evaluate the removal performance and image quality of ControlFill, where all masked regions in (\ref{eqn_ternary}) are set to $-1$. I compare the following algorithms:

\begin{itemize}
\item \textbf{LaMa} \cite{zhao2021large}: A widely used GAN-based inpainting algorithm. I compare ControlFill with LaMa qualitatively, as most of LaMa's results are blurry and not directly comparable to diffusion-based networks.
\item \textbf{Stable Diffusion XL inpainting (\textit{SDXL-Inpaint})} \cite{sdxlinpainting}: A model trained from T2I SDXL using the Laion dataset. Based on the official checkpoint, I used \textit{simple background} as the positive prompt and \textit{complex object} as the negative prompt, assuming that this prompt pair helps suppress unwanted objects inside the mask.
\item \textbf{ControlFill without condition learning (\textit{Ours w/o Cond})}: A model trained only for UNet inpainting without conditional learning.
\item \textbf{ControlFill with condition learning (\textit{Ours w/ Cond})}: Fine-tuned from \textit{Ours w/o Cond} with two embeddings (removal, creation). The removal condition was used as the positive prompt and the creation condition as the negative.
\end{itemize}

I compare these algorithms using the RORD inpainting dataset \cite{sagong2022rord}, which provides counterfactual data. The dataset consists of image pairs—one with objects and one without. This allows me to measure object removal performance using Learned Perceptual Image Patch Similarity (LPIPS) \cite{zhang2018unreasonable} and FID \cite{heusel2017gans}, comparing generated images to the counterfactual set. I randomly selected 100 images for comparison.

Additionally, I assess object removal effectiveness using the depth of the generated image \cite{ranftl2020towards}. I calculate the average depth difference inside the mask compared to the ground truth (GT) image, denoted as DEPTH DIFF. A lower value indicates better depth consistency. Since effective object removal should increase depth values inside the mask, I also measure the ratio of mean depth values between the masked and unmasked regions, denoted as DEPTH REL.
The quantitative results are summarized in Table \ref{table:removal}.

\begin{table*}[t]
\centering
\begin{tabular}{@{}lcccc@{}}
\toprule \midrule
                 & LPIPS ↓  & FID ↓    & DEPTH\_REL ↑ & DEPTH\_DIFF ↓ \\ \midrule
SDXL\_Inpainting & 0.122 & 58.873 & 1.088     & 4.292      \\
Ours w/o Cond.   & 0.110 & 62.71  & 1.072     & 3.586      \\
\textbf{Ours w/ Cond.}    & \textbf{0.1068} & \textbf{54.2367} & \textbf{1.0692}     & \textbf{2.5349}      \\ \bottomrule
\end{tabular}
\caption{Removal performance comparison}
\label{table:removal}
\end{table*}

\begin{figure*}[h!] 
\centering
\includegraphics[width=0.92\textwidth]{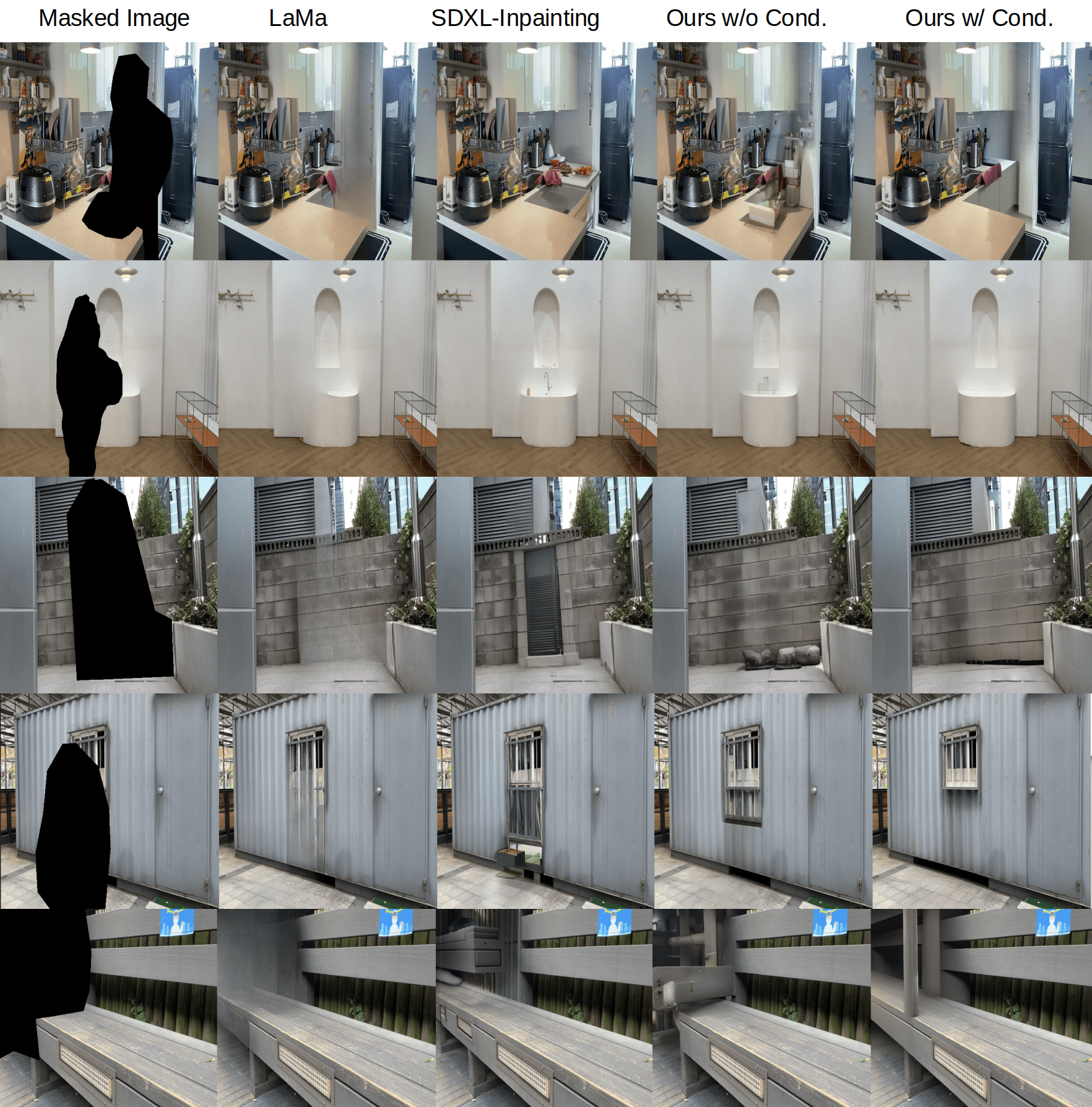}
\caption{Object removal performance comparison. LaMa produces blurry and unrealistic images due to its GAN-based inpainting approach. SDXL inpainting generates more realistic images but introduces unwanted objects inside the mask. By training both the UNet and the two conditions (removal as positive, creation as negative), ControlFill achieves clean object removal.}
\label{fig_removal_result}
\end{figure*}

\begin{figure*}[h!] 
\centering
\includegraphics[width=0.95\textwidth]{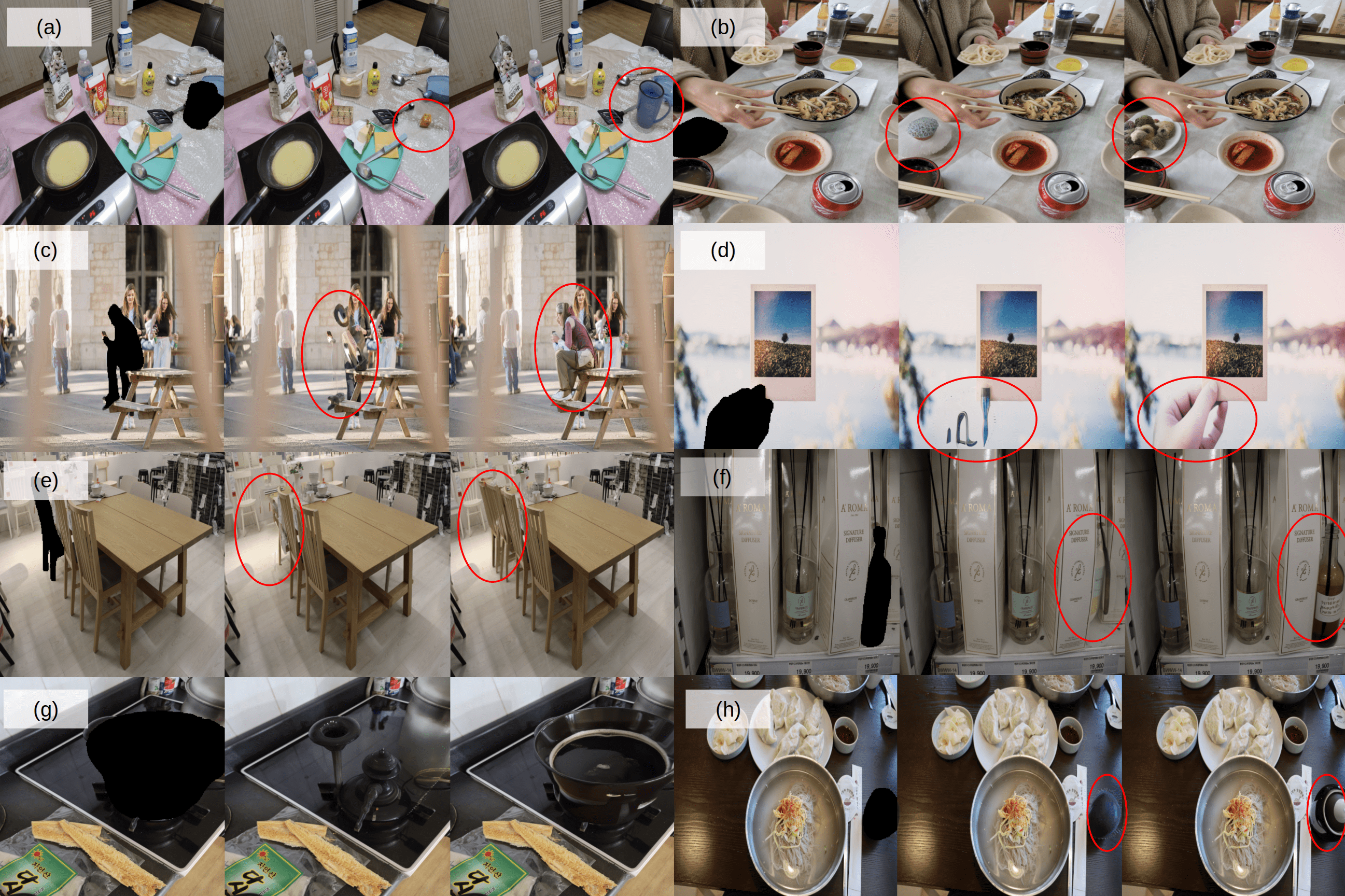}
\caption{Object creation performance. In contrast to the removal conditional, the proposed ControlFill actively creates relevant objects in the masked regions when the learned prompts \textit{creation} was used. 
Comparison with SDXL inpainting with fixed prompts (positive: objects, negative: empty background). In each triplet images, the second column was generated by SDXL, while the last column by our algorithm. ControlFill fills the masks with more consistent objects observing the context of the unmasked region.
}
\label{fig_creation_resul}
\end{figure*}

\subsection{Creation and Control Performance}

Next, I evaluate how ControlFill generates plausible content inside the mask when using the creation condition $y_{c}$ without explicit text prompts. The goal of inpainting algorithms is to actively generate relevant objects rather than simply extending the background. I assess whether the generated objects integrate well with the unmasked regions.

I collected 100 natural images and applied randomly assigned object-shaped masks. The model had no access to the masked regions' contents.

The results are shown in Figure \ref{fig_creation_resul}. As illustrated in the first row, ControlFill effectively inpainted the masks with relevant objects, such as food on plates. The model also leveraged mask shapes to generate meaningful objects, as seen in cases (c) and (d). Additionally, it successfully extended repetitive patterns, as shown in (e) and (f).

For comparison, I tested SDXL inpainting using fixed prompts (\textit{objects} as the positive prompt and \textit{empty background} as the negative). Despite using explicit prompts, SDXL failed to match ControlFill's level of creative object generation (Figure \ref{fig_creation_resul}).

Quantitatively, I compared SDXL with ControlFill using Human Preference Score V2 (HPSv2) \cite{wu2023human} and CLIP score \cite{hessel2021clipscore}, as summarized in Table \ref{table:creation}. The results confirm that ControlFill actively fills masks while considering the surrounding context.

\begin{table}[]
\centering
\begin{tabular}{lcc}
\toprule
                         & HPSv2 score  & CLIP score \\ \hline
SDXL         & 0.228 & 26.29     \\
Ours w/ Cond. & \textbf{0.232} & \textbf{26.59}    \\
\bottomrule
\end{tabular}
\caption{Creation performance comparison}
\label{table:creation}
\end{table}

\section{Conclusions}
In this paper, I confirmed the effectiveness of prompt optimization in reflecting user intentions without relying on a text encoder during the inference phase. This approach significantly reduces memory and computational requirements. Additionally, I demonstrated how user intentions can be adjusted along the spatial axis by leveraging the properties of the variational autoencoder. The proposed method effectively balances computational efficiency and user experience, making it well-suited for on-device applications.

While the generated algorithms successfully reflect user intentions without text encoders, they lack precise control over the details of the generated content. For example, in creation mode, it is not possible to specify the exact object classes to be generated inside the mask. To address this limitation, I propose class-specific prompt learning as a future direction to enhance the method’s extensibility.

\section*{Disclaimer}
This report is intended for research purposes only and does not contain any information related to commercialized products or services. All experiments and findings presented are based on research methodologies and publicly available datasets, without reference to proprietary or confidential technologies.

\bibliographystyle{unsrt}  
\bibliography{sample}

\end{document}